\documentclass[journal,twoside,web]{ieeecolor}
\usepackage{tmi}
\usepackage{cite}
\usepackage{amsmath,amssymb,amsfonts}
\usepackage{algorithmic}
\usepackage[ruled,vlined]{algorithm2e}
\usepackage{graphicx}
\usepackage{textcomp}
\usepackage{multirow}
\usepackage{multicol}
\usepackage{bigstrut}
\usepackage[normalem]{ulem}
\useunder{\uline}{\ul}{}
\usepackage{hyperref}
\hypersetup{
	colorlinks=true,
	linkcolor=blue,
	filecolor=blue,      
	citecolor=blue,
}
\usepackage{color}
\usepackage{colortbl}
\usepackage{booktabs}
\definecolor{tabcolor}{rgb}{.494,.494,.494}

\usepackage{etoolbox}
\makeatletter
\@ifundefined{color@begingroup}%
{\let\color@begingroup\relax
\let\color@endgroup\relax}{}%
\def\fix@ieeecolor@hbox#1{%
\hbox{\color@begingroup#1\color@endgroup}}
\patchcmd\@makecaption{\hbox}{\fix@ieeecolor@hbox}{}{\FAILED}
\patchcmd\@makecaption{\hbox}{\fix@ieeecolor@hbox}{}{\FAILED}

\def\BibTeX{{\rm B\kern-.05em{\sc i\kern-.025em b}\kern-.08em
    T\kern-.1667em\lower.7ex\hbox{E}\kern-.125emX}}
\begin{document}
\title{AC-Norm: Effective Tuning for Medical Image Analysis via Affine Collaborative Normalization}
\author{Chuyan Zhang, Yuncheng Yang, Hao Zheng, Yun Gu, \IEEEmembership{Member, IEEE}
\thanks{C. Zhang, Y. Yang and Y. Gu are with the Institute of Medical Robotics, Shanghai Jiao Tong University,
Shanghai 200240, China. (Email:\url{{zhangchuyan,Yaphabates,geron762}@sjtu.edu.cn}) }
\thanks{H. Zheng is with Tencent Jarvas Lab, Shenzhen, China.(Email:\url{howzheng@tencent.com})}
}

\maketitle

\begin{abstract}
Driven by the latest trend towards self-supervised learning (SSL), the paradigm of ``pretraining-then-finetuning'' has been extensively explored to enhance the performance of clinical applications with limited annotations. Previous literature on model finetuning has mainly focused on regularization terms and specific policy models, while the misalignment of channels between source and target models has not received sufficient attention.
In this work, we revisited the dynamics of batch normalization (BN) layers and observed that the trainable affine parameters of BN serve as sensitive indicators of domain information.  Therefore, Affine Collaborative Normalization (AC-Norm) is proposed for finetuning, which dynamically recalibrates the channels in the target model according to the cross-domain channel-wise correlations without adding extra parameters. Based on a single-step backpropagation, AC-Norm can also be utilized to measure the transferability of pretrained models.
We evaluated AC-Norm against the vanilla finetuning and state-of-the-art fine-tuning methods on transferring diverse pretrained models to the diabetic retinopathy grade classification, retinal vessel segmentation, CT lung nodule segmentation/classification, CT liver-tumor segmentation and MRI cardiac segmentation tasks.
Extensive experiments demonstrate that AC-Norm unanimously outperforms the vanilla finetuning by up to 4\% improvement, even under significant domain shifts where the state-of-the-art methods bring no gains. We also prove the capability of AC-Norm in fast transferability estimation.
Our code is available at \url{https://github.com/EndoluminalSurgicalVision-IMR/ACNorm}.

\end{abstract}


\section{Introduction}
\label{sec:introduction}
\IEEEPARstart{L}{earning} with few labeled data has been a longstanding problem for medical imaging analysis based on deep learning. Transfer learning (TL) has been a de-facto method to reduce the burden of labour-intensive annotations for target tasks~\cite{hosseinzadeh2021systematic, kora2022transfer} by transferring the knowledge learned from a source task to a target task. As an effective paradigm for TL, the ``pretraining-then-finetuning'' (PT-FT) workflow has been widely applied to improve performance across various medical imaging applications~\cite{lee2019explainable, choudhary2021chest}, where a network is first pretrained on a source dataset and then fintuned on a labelled target dataset with a specific task head attached to the backbone. 
Recent efforts have been dedicated to the designs of pretraining algorithms, including the fully-supervised learning (FSL)~\cite{wen2021rethinking, karimi2021transfer, zoetmulder2022domain} and the self-supervised learning (SSL) ~\cite{zhu2020rubik, taleb20203d, zhou2021models, zhou2021preservational, zhang2023dive}. 

However, the process of transfer has not received sufficient attention in the medical imaging community. The first problem is the lack of metrics for selecting a good pretrained model tailored to a specific target task. The typical practice is based on heavy experiments of finetuning on the target data. Although recent works have proposed the metrics~\cite{nguyen2020leep, you2021logme} for transferability estimation, they only adopted the static representations from the fixed pretrained model. Regarding the realistic transfer of a pretrained model to a target task, it is essential to utilize the adapted representations during finetuning for at least one complete epoch. The one-epoch backpropagation ensures that the model processes all the target data within the distribution, thereby enabling a more comprehensive evaluation of the transfer performance in a task-specific context.

\begin{figure}[!t]
\centering
\includegraphics[width=0.9\linewidth]{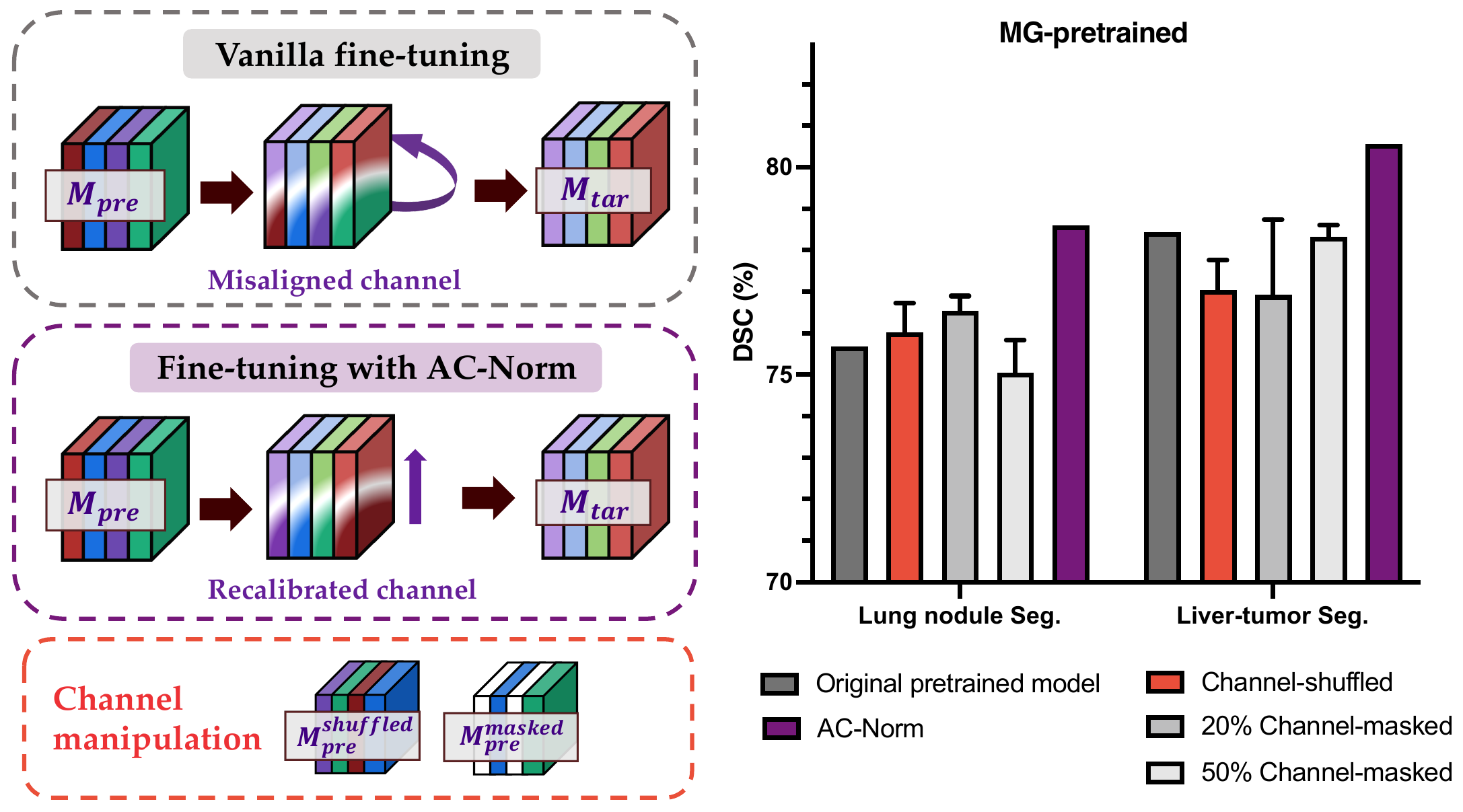}
\caption{Left: Illustration of the channel misalignment issue during transferring a pretrained model $M_{pre}$ to a target model $M_{tar}$. In the vanilla finetuning, the corresponding channels in $M_{pre}$ and $M_{tar}$ might be misaligned, while our AC-Norm recalibrates the target channels. Right: The transfer performance of original $M_{pre}$, channel-shuffled $M_{pre}$ with disrupted weight channels and channel-masked $M_{pre}$ with different ratios (the weights in masked channels are replaced with random initialization values) in Lung nodule~\cite{LUNA16} and Liver-tumor segmentation~\cite{LiTS} tasks. For all target tasks, $M_{pre}$ is obtained by the Model Genesis pretext task~\cite{zhou2021models} on LUNA16 dataset~\cite{LUNA16}. The channel-shuffling and masking operations are randomly repeated three times.
Both the channel-shuffled and channel-masked models achieve comparable performance to the original pretrained model, confirming the bottleneck of channel misalignment and channel redundancy for knowledge transfer. Our AC-Norm improves the transfer performance in both target tasks. }
\label{fig::shuffle-channel}
\end{figure}

Another under-explored problem of TL is the effective transfer of pretrained knowledge to target tasks. 
Most of the TL works initialize the target model with the corresponding pretrained weights and then completely/partially update the weights~\cite{kora2022transfer,choudhary2021chest}. 
This standard fine-tuning leads to a potential issue. 
{{As recent works~\cite{wang2019tn,huang2023reciprocal} point out, treating all channels equally during domain adaptation is sub-optimal as some source channels\footnote{The terms ``source channel'' and ``target channel'' refer to the channels in the source/pretrained model and target model respectively.} are more transferable when they capture the common patterns for the target task. Moreover, shared patterns in different domains could exist in non-corresponding channels, which impairs the transfer of source knowledge encoded in these channels to the related target channels. Intuitively, we hypothesize that such channel misalignment also poses obstacles in a general TL scenario (i.e. PT-FT).}} 
To validate the hypothesis above, we conduct a preliminary experiment as shown in Fig.\ref{fig::shuffle-channel}, where we randomly mask and shuffle the source channels in the pretrained model before initializing the target model. 
Intriguingly, both the channel-masked and channel-shuffled pretrained models attain comparable performance with the original pretrained model, suggesting two insights: (1) 
not all source channels are productive for target tasks. (2) the transferable patterns across domains are not inherently aligned along original channels. Actually, (2) is a causal factor of (1), as a mismatched channel leads to low-profit transfer.
Therefore, the key to breaking through the performance bottleneck of TL methods is to emphasize the most transferable source channels and calibrate target channels to match each to the most relevant source channel.

Thus far, finetuning approaches, which consider channel-wise transferability, have not been fully explored in the medical domain. In the general computer vision community, policy-network-based methods show potential to handle the channel misalignment issue by adaptively selecting reusable layers \cite{guo2019spottune} or filters \cite{guo2020adafilter, liu2021transtailor}. However, the layer-wise selection requires complex structures and heavy computations, conflicting with costly 3D medical tasks. Unlike these methods, we are motivated to peruse a \textbf{parameter-free} finetuning method to align the source-target channel.

Alternatively, the normalization techniques are used to align source-target feature distributions with minor modifications in unsupervised domain adaptation (UDA)~\cite{maria2017autodial, you2021alphabn, wang2019tn}. However, their setup is quite different from the standard PT-FT pipeline: (1) The source task and target task must be the same in UDA, but they are most likely to be distinct in PT-FT. This means that in PT-FT, in addition to the significant data gap, the task gap also contributes to the domain gap. (2) These normalization techniques require access to the original source data when training the target model. In PT-FT, however, the source data is unseen after pretraining.

In this paper, the transferability assessment and source-target channel misalignment are implemented with a unified framework. We first investigate the dynamics of different network modules during finetuning and uncover that the BN affine parameters are more sensitive to domain shifts than Convolutional (Conv) filters and BN statistics.
Inspired by these insights, we propose to use BN affine parameters as an indicator of domain information and devise a normalization technique called Affine Collaborative Normalization (AC-Norm) for channel-calibrated fine-tuning. AC-Norm also produces a byproduct, AC-Corr, to allow transferability estimation after a fast adaptation process of learning target-specific representations.

Our main contributions are as follows:
\begin{enumerate}
\item We propose AC-Norm, a plug-and-play and parameter-free module, to perform transferability attentive finetuning, which recalibrates target BN channels by cross-channel collaboration. 

\item Based on AC-Norm, we present an AC-Correlation (AC-Corr) metric for fast transferability estimation on various pretrained models, which for the first time considers the dynamic representations during finetuning.

\item We extensively validate the generality and superior performance of our method under 38 TL scenarios, covering both classification and segmentation target tasks with 2D and 3D models for various medical imaging modalities.
We further demonstrate the effectiveness of AC-Corr in the transferability assessment of pretrained model zoos.
\end{enumerate}
\section{Related work}

\subsection{Pretraining methods}
Driven by the desire to reduce manual annotations of medical data in source domain , Self-Supervised Learning (SSL) is progressively overtaking Fully-Supervised Learning (FSL) as the most prominent pretraining method. SSL aims to learn salient representations from the unlabelled data itself, by means of predicting the spatial relationship of patches~\cite{taleb20203d,zhu2020rubik}, pixel-level reconstruction~\cite{zhou2021models, zhou2021preservational} or contrastive learning~\cite{chen2020simple, grill2020bootstrap, zhou2021preservational}. Our work considers a more broadly applicable setup of transfer learning, in which the pretraining manners comprise both FSL and SSL.

\subsection{Finetuning methods}
In the context of transfer learning, finetuning is an essential step in adapting pretrained models to target tasks, particularly in cases where the target task differs from the source task (i.e. inductive transfer learning). Recent works on finetuning can be divided into three main categories, as detailed below.

\textbf{Regularization Tuning:}
To prevent the catastrophic forgetting, which is the tendency of the model to lose previously learnt knowledge and overfit limited target data, some regularization terms have been proposed to constrain the finetuned model to be close to the pretrained model. LI \emph{et. al}~\cite{xuhong2018explicit} and Li \emph{et. al}
~\cite{li2019delta} regularized the fintuned model with the weights or features of the pretrained model.
Chen \emph{et. al} \cite{chen2019bss} found that the features with small eigenvalues lead to negative transfer, for which they presented BSS to penalize small eigenvalues during finetuning. In the field of continual learning, Reiss \emph{et. al}~\cite{reiss2021panda} proposed PANDA-EWC (elastic weight consolidation) to resolve the dilemma of feature collapse by regularization of the change in the weights of the feature extractor. They estimated the importance of each weight to the pretraining loss and used a weighted regularization. Unfortunately, these methods are inadequate for medical applications. As claimed in \cite{you2020co}, regularization-based methods are often brittle to large domain shifts, whereas the distribution of medical images varies greatly due to many factors, such as different imaging modalities, anatomical regions and acquisition protocols. 

\textbf{Policy-network-based Tuning:}
Another line of work aims to adaptively select reusable layers~\cite{guo2019spottune} or filters~\cite{guo2020adafilter} through additional policy network components. For instance, Guo \emph{et. al}~\cite{guo2019spottune} constructed a policy network to decide whether to feed a given image into the fintuned layer or the pretrained layer. Liu \emph{et. al}~\cite{liu2021transtailor} searched an optimal sub-model for a specific target task based on a set of trainable factors measuring the target-aware filter importance.
Despite the outstanding performance of these fine-grained selective finetuning methods, they suffer from complex structures and heavy computations, conflicting with costly 3D medical tasks.

\textbf{Source-accessible Tuning:}
To further boost the final performance, source data has been borrowed during finetuning. 
You \emph{et. al}~\cite{you2020co} proposed to learn the relationship between source and target labels. Liu \emph{et. al}~\cite{liu2022improved} selected a subset of pretraining data similar to the target data, and performed semi-supervised learning with the unlabelled source data and labelled target data jointly. In medical imaging, the need for source data is constrained by ethical restrictions.

Towards effective TL across general medical imaging tasks, our goal is to explore a domain-gap-robust, low-cost and source-free finetuning method.
\subsection{Normalization in TL}
As a crucial component in deep neural networks, Batch Normalization (BN) has been proven to stabilize and accelerate training \cite{santurkar2018does}. Recently, some researchers have found that affine transformations in the BN layers endow the model with incredible expressive capabilities, no matter when training a randomly initialized model~\cite{frankle2020training} or finetuning a pretrained model~\cite{kanavati2021partial}.
In domain adaptation, designing distribution-specific normalization layers are prevalent choices to mitigate the domain gap with minor modifications.
Chang \emph{et.al.}~\cite{chang2019domain} used separate BN layers for each domain to distinguish domain-specific and domain-invariant information. In specific, separate BN statistics are used
to align different domains~\cite{you2021alphabn, maria2017autodial, wang2019tn}.
However, these methods are based on a strong assumption that the source task and the target task share the same label space, which is not satisfied under the broader TL settings. Besides, source data could not be accessed during training due to privacy protection, making it impossible to obtain source BN statistics. 

To alleviate overfitting in finetuning, Kou \emph{et.al.}~\cite{kou2020sn} introduced StochNorm to incorporate the moving and batch statistics. Inspired by recent highlights on BN affine parameters during TL~\cite{kanavati2021partial, yazdanpanah2022revisiting}, we explore the dynamics of BN affine parameters with other network parameters in various TL settings and discover that affine parameters are good indicators for domain distributions. Our method differs from prior normalization methods in two key aspects. On one side, we focus on aligning domains by BN affine parameters rather than BN statistics, which also circumvents the demand for source data. On the other side, we first propose to resolve the issue of channel misalignment during finetuning.
\subsection{Transferability Estimation}
In recent years, some transferability metrics have been proposed to facilitate the fast selection of pretrained models~\cite{nguyen2020leep, you2021logme,  huang2022frustratingly}.
The target images are first 
projected into the source embedding space of a given pretrained model.
Then, the transferability of the pretrained model can be measured by the relationship between the source-embedded target features and the target task.
The formulation of the relationship is one of the heated research interests. LEEP~\cite{nguyen2020leep} constructs the joint distribution over pretrained labels and the target labels, and uses an empirical predictor to assess the transferability. LogME~\cite{you2021logme} estimates the maximum value of label evidence given source-embedded target features. 
TransRate~\cite{huang2022frustratingly} computes the mutual information of the source-embedded target features and target labels. 
These metrics are calculated directly from the static representations of the fixed pretrained model without training on the target data. This paper first explores measuring the transferability based on the adapted representations during one-epoch finetuning.

\section{Methodology}
In this section, we first introduce the problem setting of transfer learning from a zoo of models. Then we investigate the dynamics of network components during finetuning. Based on the observations, we introduce our AC-Norm approach, which consists of channel alignment and adaptive tuning. We further propose an efficient TE metric.

\subsection{Preliminaries}
In TL, let $D_s$ and $D_t$ respectively denote the source dataset and target dataset. We are first given a pretrained model $M_0$ (the optimal source model $M_s$), which is optimized with $D_s$ by minimizing the loss function $L_s$ of a source task $T_s$:
\begin{equation}       
M_0 = \arg \min_{M_s}L_s(D_s;M_s, T_s)
\end{equation}
To eliminate the need for manual annotations of $D_s$ in medical imaging, self-supervised $T_s$ is preferred over traditional supervised learning. $M_0 $ is commonly split into two parts including a general feature extractor $G_0$ and a task-specific head $H_0$. 

For a specific task $T_t$, we construct the target model $M_t=\{G_t, H_t\}$, where $G_t$ is initialized by $G_0$ and $H_t$ is randomly initialized. The architectures of $G_t$ and $G_0$ are identical, whereas the task head $H_t$ might differ from $H_s$.
The finetuning stage aims to find an optimal target model $M^*_t$ on $D_t$ 
by minimizing the loss function $L_t$ of $T_t$, where the functional space of $G_0$ provides a good starting point for the optimization process:
\begin{equation}       
M^*_t = \arg \min_{M_t}L_t(D_t; G_{0\rightarrow t}, H_t, T_t)
\end{equation}

For domain adaptation with $T_s=T_t$, only the distribution shift between $D_s$ and $D_t$ is considered. In inductive TL, however, useful knowledge transfer in $G_{0\rightarrow t}$ is hampered by both the task gap of $T_s$ and $T_t$ and the data gap of $D_s$ and $D_t$, as illustrated in Fig.\ref{fig::acn-overview}~(A). Hence, finetuning techniques are imperative to effectively reuse the general representations from $G_0$. 

\textbf{Regularization Tuning.} To prevent $M^*_t$ from overfitting to $D_t$, some regularizes $\Omega$ \cite{xuhong2018explicit,li2019delta,chen2019bss} are added to the optimization of $M_t$:
\begin{equation}       
M^*_t = \arg \min_{M_t}L_t(D_t; G_{0\rightarrow t}, H_t, T_t) + \Omega(G_0, G_{0\rightarrow t})
\end{equation}

\textbf{Importance-aware Tuning.} More recent work achieves importance-aware fine-tuning through selectively transferring $G_0$ to $G_t$. In specific, the optimization of $G_t$ is guided by layer-wise or filter-wise importance values based on additional policy networks \cite{guo2019spottune, liu2021transtailor}:
\begin{equation}       
M^*_t = \arg \min_{M_t}L_t(D_t;A_{\theta}(G_t;G_0), H_t, T_t)
\end{equation}
in which $A_{\theta}$ represent an adapter that selectively reuses $G_0$ for $G_t$. Existing solutions to $A$ require extra parameters $\theta$, leading to a high computational burden. Different from these works, the proposed method in this paper introduced no additional parameters.
\begin{figure}[]
\centering
\includegraphics[width=0.9\linewidth]{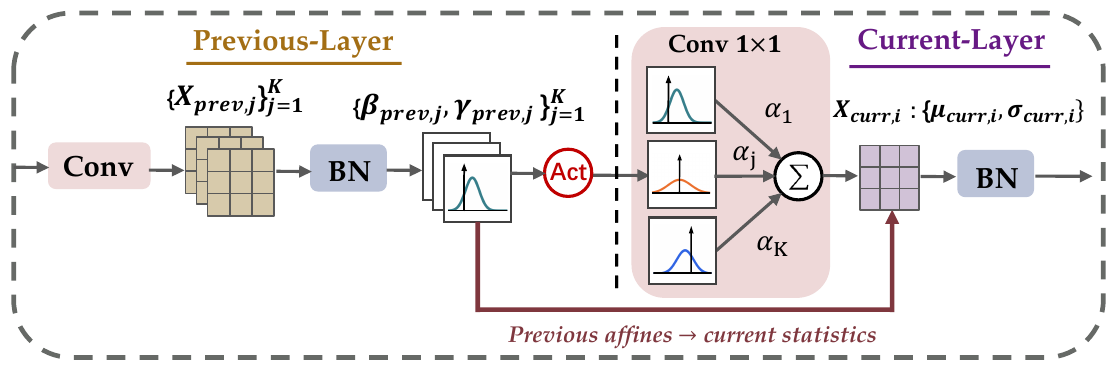}
\caption{Flow diagram of a previous layer and current layer. Assuming the convolutional kernel size is 1, the BN statistics in the current layer depend on the BN affine parameters in the previous layer.}
\label{fig::bn_flow}
\end{figure}

\subsection{BN Formulations}\label{sec::bn_formulation}
In the standard BN layer~\cite{ioffe2015batch}, let $K$ denote the number of channels and $N$ denote the product of mini-batch size $M$ and spatial dimension ($D\times H\times W$ for 3D images and $H\times W$ for 2D images). The input features $x \in R^{N\times K}$ are first standardized by channel-wise mean $\mu_j$ and standard deviation $\sigma_j$:
$\hat{x}_{n,j}=\frac{x_{n,j}-\mu_j}{\sqrt{\sigma_j^2+\epsilon}}$, where $\epsilon$ is a small constant to avoid zero division, $n\in [1,N]$, and $j \in [1,K]$. Note that mini-batch statistics $\{\mu, \sigma\}$ and moving statistics $\{\widetilde{\mu}, \widetilde{\sigma}\}$ are used for training and inference respectively.
Along each channel, the standardized features are assumed to be in the normal distribution $N(0, 1)$ and then scaled and shifted to $N(\beta_j, \gamma^2_j)$ by learnable affine parameters: $f_{n,j}=\gamma_j{\cdot}{\hat{x}_{n,j}}+\beta_j$. 

Given an output feature map of the previous convolutional layer with $K$ channels: $\{x_{prev,j}\}^K_{j=1}$, the next BN layer normalizes each channel $x_{prev,j}$ to a Gaussian distribution $N(\beta_{prev,j}, {\gamma^2_{prev,j}})$. Then, each channel-wise Gaussian distribution is passed through an activation function, leading to rectified mean and variance: $\mu_{prev,j}=\mu_{act}\cdot\beta_{prev,j}$, ${\sigma^2}_{prev,j}=\sigma^2_{act}\cdot{\gamma^2}_{prev,j}$, as claimed in \cite{davis2022revisiting}. Next, the rectified distribution is fed into the current convolutional layer. For simplicity, we assume a $1\times1$ filter is used and ignore the bias. Then, the convolved output is the weighted sum of $K$ previous Gaussian distributions.
Let $X_{curr,i}$ denote $i$-th channel of the convolved output and $\{\alpha_j\}^K_{j=1}$ denote the filter weight. The mini-batch statistics of $x_{curr,i}$ can be obtained:
\begin{eqnarray}   \label{eq::corr_of_affine_sta}    
\begin{aligned}
&\mu_{curr,i} = \mu_{act}\sum^K_{j=1}\alpha_j\cdot\beta_{prev,j}\\ &{\sigma^2}_{curr,i} = {\sigma_{act}}^2\sum^K_{j=1}{\alpha_j}^2\cdot{\gamma^2}_{prev,j}
\end{aligned}
\end{eqnarray}

Eq.\eqref{eq::corr_of_affine_sta} indicates that the affine parameters in the previous BN layer determine the mini-batch statistics in the current BN layer. According to the chain rule, the affine parameters are linked with the mini-batch statistics via gradient propagation. When optimizing a network, historical mini-batch statistics constitute the moving statistics and affine parameters stand for the current optimal statistics after gradient accumulation. Therefore, we can regard the running statistics as temporal-momentum statistics and the affine parameters as gradient-accumulated statistics.

\subsection{Dynamics of Layers}
\begin{figure}[]
\centering
\includegraphics[width=0.8\linewidth]{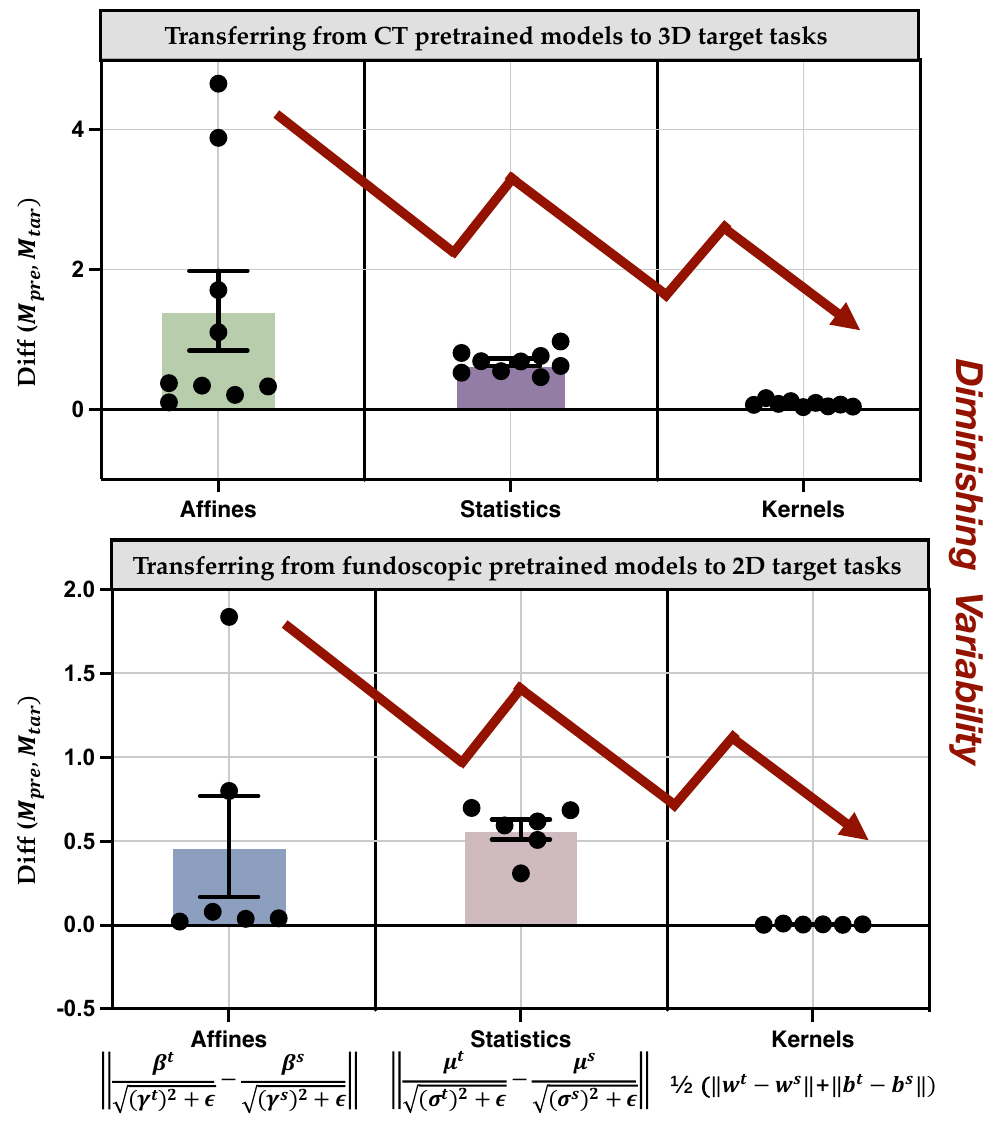}
\caption{The updating degree of different network components in finetuning the pretrained model $M_{pre}$ to the target model $M_{tar}$ across various TL scenarios. In particular, we transfer 3 CT pretrained models (MG/RPL/SimCLR) to 3 target tasks (NCC/NCS/LCS) and 3 fundoscopic pretrained models (MG/RPL/SimCLR) to 2 target tasks (DFC/VFS).The details of all datasets and tasks are presented in Section~\ref{sec::datasets}. }
\label{fig::affine_sta_kernel}
\end{figure}
\begin{figure*}[!]
\centering
\includegraphics[width=0.8\linewidth]{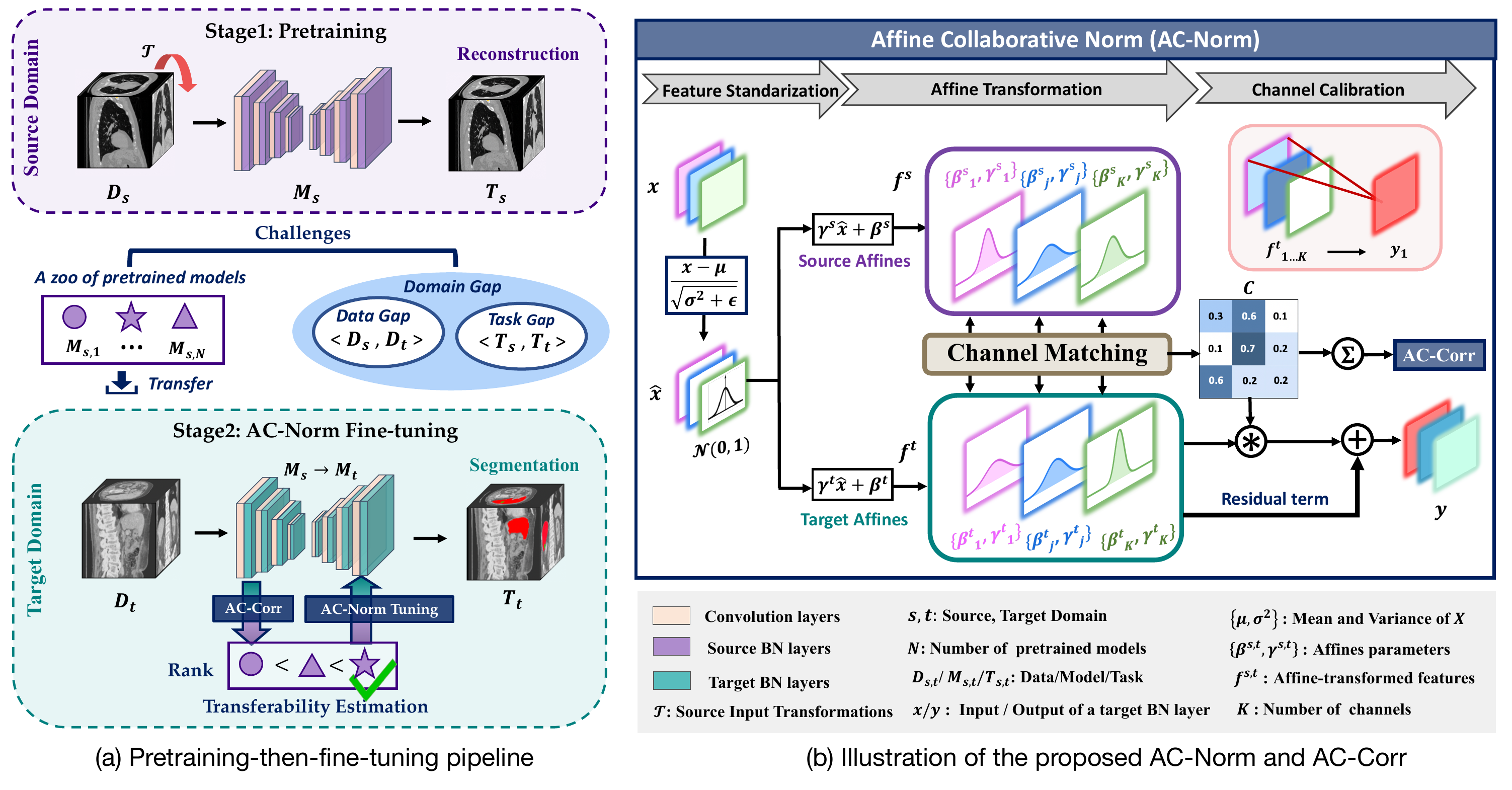}
\caption{The overview of the proposed AC-Norm for improving the transfer performance and AC-Corr for transferability estimation. (a) The pipeline of pretraining and finetuning: The challenges in achieving efficient TL consist of model selection among numerous pretrained models and notable domain gaps. A good pretrained model is first selected via our AC-Corr. Then, plugging AC-Norm into the target model during the finetuning phase can boost the transfer efficiency. (b) The illustration of AC-Norm: the cross-channel transferability is estimated based on source and target affines and a channel calibration module is constructed to draw more attention of the target model to transferable channels.}
\label{fig::acn-overview}
\end{figure*}

The BN layer has been considered to be a crucial component in bridging the distribution gap between the source and target datasets. Li \emph{et al.} \cite{li2018adaptive} assumed that the weight matrixes always carry the label information while the BN statistics store the domain-related knowledge. This interpretation leads to questions about the role of the BN affine parameters during TL. A few pilot works have distinguished BN affine parameters from other weight layers in few-shot TL \cite{yazdanpanah2022revisiting} or traditional supervised TL \cite{kanavati2021partial}. However, the dynamics of affine parameters in a wide PT-FT workflow toward medical imaging tasks remain to be explored.
In Fig.\ref{fig::affine_sta_kernel}, we collect the average differences of encoder layers between the pretrained and the fintuned model in terms of affine weights($|\frac{\beta}{\sqrt{(\gamma)^2+\epsilon}}, \epsilon=1e^{-5}$) and moving statistics ($\frac{\mu}{\sqrt{\sigma^2+\epsilon}}$) of BN layers as well as the kernel weights ({$w, b$}) of convolutional layers. It is worth noting that the difference of a component represents its
reusability during transfer. We observe that the updated magnitude of affines varies dramatically with different source and target tasks while kernel weights are slightly updated. This indicates that kernel weights with high reusability
contain more domain-generic information while affine weights with volatile reusability tend to be more domain-specific. Therefore, we propose to use BN affines as the domain indicator rather than commonly used BN statistics. 

\subsection{Affine Collaborative Normalization}
Based on the aforementioned observations, we propose the Affine Collaborative Normalization (AC-Norm) for efficient finetuning, as illustrated in Fig.~\ref{fig::acn-overview}(b).
Formally, ley $\{\gamma^s,\beta^s\} \in R^{K}$ denote the source affines in the pretrained model and $\{\gamma^t,\beta^t\} \in R^{K}$ (initialized by source affines) denote the trainable target affines. According to  Section~\ref{sec::bn_formulation}, the standardized features $\hat{x} \in R^{N\times K}$ can be transformed to the source feature space through source affines ($f^s$) while to the target feature space via target affines ($f^t$). Under the basic assumptions of BN, the channel transferability can be quantified by the negative discrepancy between source and target distributions:\\
\begin{equation}
C_{p,q} = \frac{exp({-|z^t_{p}-z^s_{q}|/t)}}{\sum_{j=1}^K({exp(-|z^t_{p}-z^s_{j}|/t))}}, p,q \in [1, K]
\end{equation}
where $z^{s, t} = \frac{\beta^{s, t}}{\sqrt{({\gamma^{s,t}})^2+\epsilon}}$ describes the distributions of affine-transformed features and $t$ is a temperature term to adjust the distribution of important values.
Other alternatives for $z$ are analyzed in the ablation study. Though $z^t_{p}$ is initialized by $z^s_{p}$, the weight updating degree during finetuning varies with the reusability. The $p$-th row in $C \in R^{K\times{K}}$ reflects the correlation from all source channels to the $p$-th target channel. 
To prevent the negative transfer, we focus on the non-corresponding channel only when its importance is greater than that of the corresponding channel, resulting in a sparse $C$:
\begin{equation}\label{eq::sparsity}
C_{p, q}=\left\{
	\begin{aligned}
	C_{p, q} \quad  if~ C_{p, q} \geq C_{p, p}\\
	{0\indent\indent} \quad otherwise\\
	\end{aligned}
	\right.
\end{equation}

The Affine Collaborative matrix $C$ can facilitate TL from two perspectives: guiding the channel recalibration during finetuning and assessing the transferability of the pretrained model to the target task.

1) \emph{Channel Recalibration}: With the transferability measured in Eq.~\ref{eq::sparsity}, our intuition is that a channel with high transferability indicates it needs to be assigned more attention values in the target model and vice versa. More critically, a target channel might need to pay attention to another target channel that inherits from the associated source channel.
Therefore, we recalibrate the target BN channels to realize the importance-aware finetuning based on the channel collaborations in $C$: \\
\begin{equation}
\gamma^{c} = C[\gamma^t_1,..., \gamma^t_K]^T , \beta^{c} = C[\beta^t_1,..., \beta^t_K]^T
\end{equation}
Finally, the output feature $y$ is obtained by:
\begin{equation}
y_{n, j} = f^t_{n, j} + (\gamma^{c}_j{\cdot}\hat{x}_{n, j} +  \beta^{c}_j), n \in [1, N], j \in [1, K]
\end{equation}

To avoid the loss of information, we reserve the original features $f^t$ and form the calibrated features as a residual term. Algorithm~\ref{algorithm1} presents the pseudo-code of AC-Norm.

2) \emph{Transferability Estimation (TE)}: Since $C$ in AC-Norm represents the relatedness of source and target features, AC-Norm is endowed with a function of transferability estimation. Concretely, we sum the channel-wise transferability values at a layer $l$:
\begin{equation}
AC\text{-}Corr = \sum_{p=1}^{K}\sum_{q=1}^{K}C^l_{p, q}
\end{equation}
In practice, we fintune a pretrained model with AC-Norm for only one epoch and compute AC\text{-}Corr of the last layer in the encoder to assess the transferability of a pretrained model.

\begin{algorithm}[!t]
	\caption{Affine collaborative Normalization (AC-Norm).}
\label{alg:algorithm1}
\algsetup{linenosize=\tiny} \footnotesize
\KwIn{mini-batch features $x=\{x_{n, j}\}, n\in[1, N], j\in[1, K]$; source affine parameters from a pretrained model $\beta^s, \gamma^s$ for $K$ channels; trainable target affine parameters $\beta^t, \gamma^t$ (initialized by $\beta^s, \gamma^s$);  a temperature hyper-parameter $t$}
\KwOut{y=AC-Norm(x).}  
\textbf{Training:}\\
mini-batch statistics:\\
   $\mu_j \gets \frac{1}{N}\sum_{n=1}^{N}{x_{n,j}}$, ${\sigma_j}^2 \gets \frac{1}{N}\sum_{n=1}^{N}{(x_{n,j}-\mu_j)^2}$\\
   feature standardization:\\
   $\hat{x}_{n,j} \gets \frac{x_{n,j}-\mu_j}{\sqrt{{\sigma_j}^2+\epsilon}}$\\ 
   domain per-channel distributions:\\
  $z_j^s \gets \frac{\beta_j^s}{\sqrt{(\gamma_j^s)^2+\epsilon}}$, $z_j^t \gets \frac{\beta_j^t}{\sqrt{(\gamma_j^t)^2+\epsilon}}$ \\
  cross-channel transferability:\\
  $C_{p,q} \gets \frac{exp({-|z^t_{p}-z^s_{q}|/t)}}{\sum_{j=1}^K({exp(-|z^t_{p}-z^s_{j}|/t))}}, p,q\in [1, K]$\\
    $C_{p, q} = C_{p, q}~if~ C_{p, q} \geq C_{p, p} ~else~ 0 $\\ 
    target affine transformations\\
     ${f^t_{n,j}} = \gamma^t_j{\cdot}{\hat{x}_{n,j}}+ \beta^t_j$\\
     channel recalibration\\
     $\gamma^{c} \gets C[\gamma^t_1,..., \gamma^t_K]^T , \beta^{c} \gets C[\beta^t_1,..., \beta^t_K]^T$\\
   ${y_{n,j}} \gets {f^t_{n,j}} + (\gamma^{c}_j{\cdot}{\hat{x}_{n,j}} + \beta^{c}_j)$ \\
 \label{algorithm1}
 \end{algorithm}

\section{Experiments}
\subsection{Datasets and Tasks}\label{sec::datasets}
\begin{figure}[]
\centering
\includegraphics[width=0.8\linewidth]{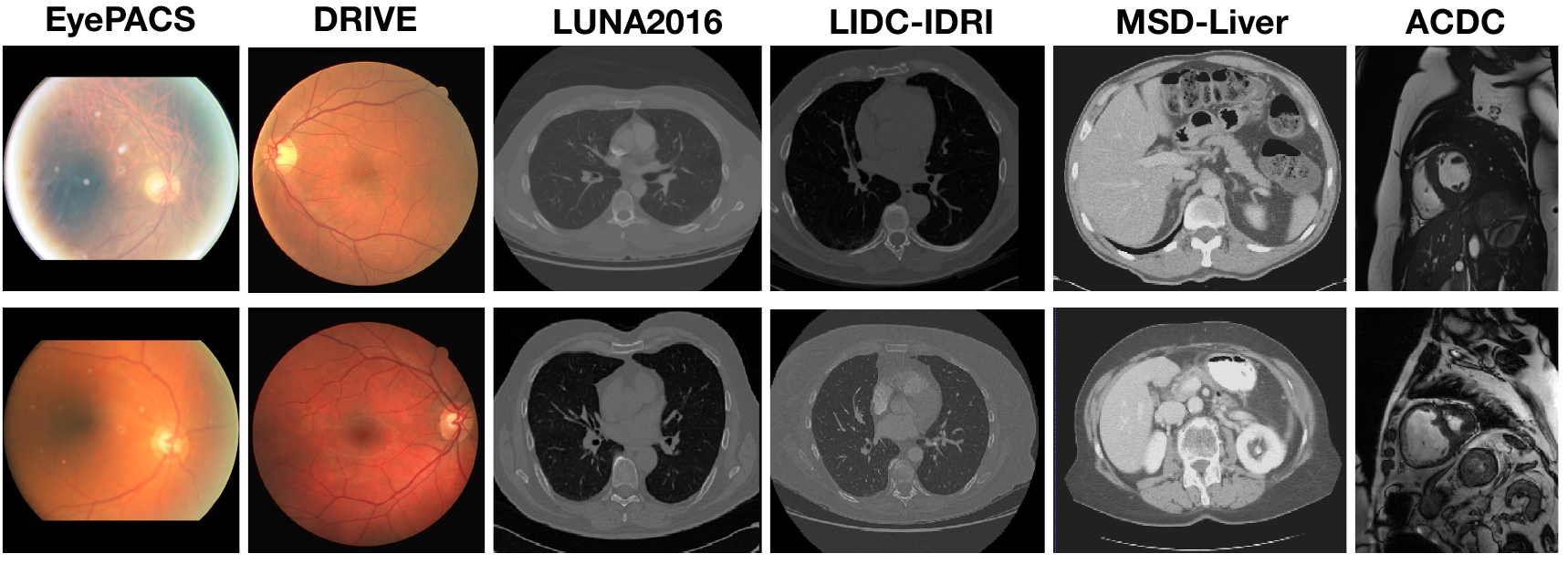}
\caption{Examples of the medical image datasets used in this paper. The CT images in LUNA2016 and LIDC-IDRI dataset are partly overlapped. The selection of datasets for transfer learning actually follows a line of recent literature~\cite{wen2021rethinking, zhou2021models, zhou2021preservational}.}
\label{fig::example_data}
\end{figure}

We conducted experiments on six medical image datasets with various modalities, including two fundus image datasets, three computed tomography (CT) image datasets, and one magnetic resonance imaging (MRI) dataset to evaluate the finetuning approaches. The example images for each dataset are exhibited in Fig.~\ref{fig::example_data}.

\textbf{Target datasets and tasks.}
 The details and setup of six target tasks are shown in Table~\ref{tab::target_dataset}. The 2D target tasks are DFC (diabetic retinopathy grade classification in EyePACS dataset~\cite{eyepacs}) and VFS (retinal vessel segmentation in DRIVE dataset~\cite{drive}). 
The 3D target tasks consist of NCS (Lung nodule segmentation in LIDC-IDRI dataset\cite{LIDC}), NCC (Lung nodule false positive reduction in LUNA2016 dataset\cite{LUNA16}) and LCS (Liver and tumor segmentation in MSD-Liver dataset)\cite{MSD, simpson2019large}, following previous works \cite{zhou2021models, zhou2021preservational} in transfer learning. Besides, we validate the proposed methods on transferring CT pretrained models to an MRI task: CMS (Cardiac segmentation on ACDC dataset~\cite{acdc}). 

\textbf{Source datasets and tasks.}
Both FSL and SSL source tasks are implemented to evaluate the performance of different pretrained models. 
{{To perform FSL pretraining, we cross-transfer the 2D or 3D target tasks.}}
For 2D SSL pretraining, we adopt
RPL~\cite{taleb20203d}, Jigsaw~\cite{taleb20203d}, PCRL~\cite{zhou2021preservational}, MG~\cite{zhou2021models}, SimCLR~\cite{chen2020simple} and BYOL~\cite{grill2020bootstrap} proxy tasks. Then, we construct a 2D pretraining dataset by selecting 28k high-quality fundus images from EyePACS datase~\cite{eyepacs} based on the gradability scores in ~\cite{voets2019reproduction}. 
For 3D SSL pretraining, RPL~\cite{taleb20203d}, RKB+ ~\cite{zhu2020rubik}, SimCLR~\cite{chen2020simple}, PCRL~\cite{zhou2021preservational} and MG~\cite{zhou2021models} are adopted as representative proxy tasks. 
The 3D pretraining dataset is composed of 623 chest CT scans from LUNA 2016~\cite{LUNA16}. Note that we keep non-overlapping between the source data with the target test data to prevent test data leakage. 
\begin{table}[]
\caption{Details of target datasets and tasks. Here we respectively use a small and large test subset for DFC. \dag The first letter stands for the object of interest (``D'' for Diabetic Retinopathy, ``N'' for lung nodule, etc); the second letter denotes the modality (``F'' for Fundoscopic, ``C'' for CT, etc); the last letter stands for the task formulation (``C''for classification, ``S''for segmentation).}\label{tab::target_dataset}
\resizebox{0.48\textwidth}{!}{
\begin{tabular}{c|cccc}
\hline
Modality                  & Dataset   & Input Size &  Task{\dag}& Train:Val:Test (case)   \\ \hline
\multirow{2}{*}{2D Fundoscopic} & EyePACS   & 512$\times$512    & DFC  & 3511:1090:10663
\\
                          & DRIVE     & 560$\times$576    & VFS  & 20:5:15                \\ \hline
\multirow{3}{*}{3D CT}    & LUNA16    & 48$\times$48$\times$48   & NCC  & 445:178:265            \\
                          & LIDC-IDRI   & 64$\times$64$\times$32   & NCS  & 510:100:408            \\
                          & MSD-Liver & 160$\times$160$\times$32 & LCS  & 100:15:15              \\ \hline
3D MRI                    & ACDC      & 320$\times$320$\times$16 & CMS  & 70:10:20               \\ \hline
\end{tabular}}
\end{table}

\begin{table*}[t]
\renewcommand{\arraystretch}{1.2}
\centering
\caption{Results of transferring the MG pretrained models to NCS, LCS, NCC, DPC and VFS target tasks. The best results are in \textbf{bold} and the second best results are \underline{underlined}}\label{tab::MG-comparison}
\resizebox{0.99\textwidth}{!}{
\begin{tabular}{cl|cc|ccc|cc|cc|ccc}
\hline
\multicolumn{2}{c|}{\multirow{2}{*}{Method}}                                & \multicolumn{2}{c|}{NCS}        & \multicolumn{3}{c|}{LCS (DSC$\uparrow$)}                         & \multicolumn{2}{c|}{NCC}                          & \multicolumn{2}{c|}{DFC}        & \multicolumn{3}{c}{VFS}         \\ \cline{3-14} 
\multicolumn{2}{c|}{}                                                       & DSC$\uparrow$         & mIoU$\uparrow$         & Liver          & Tumor           & Avg          & AUC$\uparrow$        & \multicolumn{1}{c|}{ACC$\uparrow$}         & KAPPA$\uparrow$             & ACC$\uparrow$          & CL-DSC$\uparrow$          & SEN$\uparrow$       & ACC$\uparrow$        \\ \hline
\multicolumn{1}{c|}{From-scratch}          & Random Init.              & 73.79          & 80.86          & 93.19         & 60.31          & 76.75          & {98.53}           & 97.00       & 76.96          & 77.07         & 77.28          & 75.45  &94.78       \\ \hline
\multicolumn{1}{c|}{\multirow{2}{*}{Baseline}} & MG-Vanilla-FT             & 75.69          & 82.13          & 94.03          & \underline{62.85}    & \underline{78.44}    & {98.01}        & 96.72       & 77.96          & 77.27          & 79.64          & 78.63    &95.10     \\
                       \multicolumn{1}{c|}{}     & MG-WarmUp~\cite{taleb20203d}            & 74.84          & 80.53          & 93.87          & 61.57          & 77.72          & {98.24}          & 96.55       &  \underline{78.01}         & 77.38    &       79.89         &   78.91     &\underline{95.16}          \\ \hline
\multicolumn{1}{c|}{\multirow{3}{*}{UDA method}} & + $\alpha$-BN~\cite{you2021alphabn} & 75.64          & 81.85          & 75.82          & 25.99          & 50.91          & {84.72}          & 64.38       & 67.95          & 76.17          &        \underline{79.99}        &  79.31 &95.06             \\\multicolumn{1}{c|}{} 
                           & + Auto-DIAL\cite{maria2017autodial}                & 76.27          & 82.34          & 93.91          & 49.68          & 71.80          & {97.41}          & 96.20       & 68.99          & 75.77          &    79.34            &     78.73     &94.82       \\
                        \multicolumn{1}{c|}{}    & + TransNorm\cite{wang2019tn}             & \underline{76.79}    & \underline{82.56}    & \underline{94.57}    & 53.11          & 73.84          & {99.02}          &     98.37   & 75.20          &  77.49         &        79.69        &      77.53         &94.17 \\ \hline
\multicolumn{1}{c|}{\multirow{4}{*}{FT method}}  & + StochNorm\cite{kou2020sn}           & 75.90          & 81.57          & 94.13          & 52.41          & 73.27          & {\underline{99.37}}   & \underline{98.61} & 77.82          & 75.09       &        76.47        &     \underline{79.87}       &93.98    \\
                         \multicolumn{1}{c|}{}   & + BSS\cite{chen2019bss}                    & 75.60          & 81.78          & 94.54          & 61.92          & 78.23          & 98.07         &   97.37     &  77.90      & \underline{77.69}       &        79.82        &   78.12     &94.86        \\
                      \multicolumn{1}{c|}{}    & + PANDA-EWC\cite{reiss2021panda}              & 75.12          & 81.71          & 94.53          & 60.66          & 77.60          & 98.21       & 97.48       & 76.90        & 77.17          &          79.80     &     77.62     &95.01        \\ 
                       \multicolumn{1}{c|}{}   & + AC-Norm (Ours)          & \textbf{78.60} & \textbf{82.87} & \textbf{95.14} & \textbf{65.84} & \textbf{80.56} & {\textbf{99.65}} & 
  \textbf{99.15}               & \textbf{78.75} & \textbf{77.80} & \textbf{80.87} & \textbf{80.15} & \textbf{95.24}\\ \hline
\end{tabular}}
\end{table*}

\subsection{Settings and Evaluations}
\textbf{Implementation Details.}
All experiments were run with PyTorch on two NVIDIA RTX 3090 GPUs.
For a fair comparison, we adopt the typical 2D/3D U-Net \cite{ronneberger2015unet, cciccek20163d} as the backbone for all 2D/3D source and target tasks. Especially for classification tasks, we use U-Net variants by replacing the decoder with fully connected layers. The detailed architecture and hyper-parameter settings are the same as the benchmark in ~\cite{zhang2023dive}. Note that we use a larger test set in DFC than in \cite{zhang2023dive}.
To ensure the efficiency of SimCLR on multi-object medical images, we crop each positive pair with a constraint of IoU $>$ 0.25.  In NCC, we use a mean-resampling scheme of randomly sampling the same class number at every epoch to solve the severe class imbalance problem. 

\textbf{Baselines of transfer learning.} In this work, the proposed method is compared with previous finetuning methods including: StochNorm~\cite{kou2020sn}, BSS~\cite{chen2019bss}, PANDA-EWC~\cite{reiss2021panda}. StochNorm ~\cite{kou2020sn} incorporates the running statistics and mini-batch statistics during finetuning to avoid over-reliance on some sample statistics; BSS~\cite{chen2019bss} penalizes small singular values to suppress untransferable components; PANDA-EWC~\cite{reiss2021panda} leverages elastic weight consolidation (EWC) to regularize the weight change of pretrained models and finetuning models.
To further compare with prevalent normalization methods in UDA, including $\alpha$-BN\cite{you2021alphabn}, Auto-DIAL\cite{maria2017autodial} and TransNorm\cite{wang2019tn}, the pretrained moving statistics are used as the substitute for batch statistics of source features, due to unseen source data in the standard finetuning.  In all TL settings, the vanilla finetuning serves as the baseline and all other methods are intended to improve the transfer performance over the baseline. To ensure fairness, we apply the same training hyper-parameters for all compared finetuning methods.

\textbf{Evaluation Metrics.}
Depending on the characteristics of medical imaging tasks, we select a variety of metrics to evaluate the target performance: quadratic weighted KAPPA~\cite{eyepacs} and Accuracy (ACC) for DFC task; Centerline-in-volume Dice Similarity Coefficient (CL-DSC)~\cite{paetzold2019cldice} and Sensitivity (SEN) for VFS
task; Area Under the ROC Curve (AUC) scores and ACC for NCC task; Dice Similarity Coefficient (DSC) and mean Intersection over Union (mIoU) with thresholds ranging from 0.5 to 0.95 for NCS task; DSC and 95 percentile of the Hausdorff Distance (HD95) for LCS and CMS tasks.

\subsection{Comparisons with the State-of-the-Art Methods.}
We first compare the proposed method with baselines (vanilla full finetuning and warmup finetuning~\cite{taleb20203d}), existing finetuning schemes and modified UDA normalization techniques on the transfer of the MG ~\cite{zhou2021models} pretrained models to diverse target tasks. We use 3D-MG pretrained model for NCS, LCS and NCC and 2D-MG pretrained model for DFC and VFS.
The performances are reported in Table~\ref{tab::MG-comparison}, which shows that AC-Norm consistently outperforms all other approaches in diverse target tasks.
Unfortunately, most of the compared methods fail to improve the baseline in a larger domain gap, highlighting the need for an effective finetuning strategy in the medical context.

\begin{table}[]
\centering
\caption{Quantitative comparisons in various TL settings on 2D fundoscopic datasets. The best results are in bold and the positive average gain of each FT method against the baseline is highlighted in blue.}\label{tab::results_fundus}
\resizebox{0.48\textwidth}{!}{
\begin{tabular}{p{0.6cm}<{\centering}p{1.4cm}<{\centering}|l|cc|cc}
\hline
 \multicolumn{2}{c|}{\multirow{2}{*}{PT Method}}                                                                         & \multirow{2}{*}{FT Method} & \multicolumn{2}{c|}{DFC}                                      & \multicolumn{2}{c}{VFS} \\ \cline{4-7}
\multicolumn{2}{c|}{}                                                                                                   &                            & KAPPA$\uparrow$          & ACC$\uparrow$                & CL-DSC$\uparrow$              & SEN$\uparrow$                 \\ \hline
\multicolumn{1}{c|}{\multirow{20}{*}{Self-supervised}} & \multirow{4}{*}{RPL}                                                       & Vanilla-FT                          & 77.17          & 76.25          & 78.98               & 79.05              \\
\multicolumn{1}{c|}{}                      &                                                                            & StochNorm~\cite{kou2020sn}                          & 76.80          & 77.03          & 78.98               & 77.63              \\
\multicolumn{1}{c|}{}                      &                                                                            & BSS~\cite{chen2019bss}                           & 77.25          & 76.77          & 79.05               & 76.18              \\ \cline{3-7} 
\multicolumn{1}{c|}{}                      &                                                                            & AC-Norm (Ours)                       & \textbf{78.32} & \textbf{77.18} & \textbf{80.08}      & \textbf{81.61}     \\ \cline{2-7} 
\multicolumn{1}{c|}{}                      & \multirow{4}{*}{Jigsaw}                                                    & Vanilla-FT                          & 78.07          & 76.91          & 78.57               & 77.89              \\
\multicolumn{1}{c|}{}                      &                                                                            & StochNorm~\cite{kou2020sn}                   & 75.98          & \textbf{78.14} & 77.20               & 77.21              \\
\multicolumn{1}{c|}{}                      &                                                                            & BSS~\cite{chen2019bss}                                & 76.96          & 77.24          & 78.25               & 77.19              \\ \cline{3-7} 
\multicolumn{1}{c|}{}                      &                                                                            & AC-Norm (Ours)                             & \textbf{78.89} & 77.95          & \textbf{79.47}      & \textbf{79.50}     \\ \cline{2-7} 
\multicolumn{1}{c|}{}                      & \multirow{4}{*}{PCRL}                                                      & Vanilla-FT                  & 77.76          & 76.58          & 78.98               & 76.93              \\
\multicolumn{1}{c|}{}                      &                                                                            & StochNorm~\cite{kou2020sn}                      & 77.86          & 75.46          & 77.78               & 78.90              \\
\multicolumn{1}{c|}{}                      &                                                                            & BSS~\cite{chen2019bss}                             & 77.45          & 76.88          & 78.91               & 78.27              \\ \cline{3-7} 
\multicolumn{1}{c|}{}                      &                                                                            & AC-Norm (Ours)                              & \textbf{78.81} & \textbf{77.04} & \textbf{80.08}      & \textbf{80.43}     \\ \cline{2-7} 
\multicolumn{1}{c|}{}                      & \multirow{4}{*}{SimCLR}                                                    & Vanilla-FT                        & 77.97          & 77.42          & 80.82               & 79.18              \\
\multicolumn{1}{c|}{}                      &                                                                            & StochNorm~\cite{kou2020sn}                    & 75.77          & 76.69          & 81.61               & 81.21              \\
\multicolumn{1}{c|}{}                      &                                                                            & BSS~\cite{chen2019bss}                            & 76.65          & 76.67          & 80.13               & 77.39              \\ \cline{3-7} 
\multicolumn{1}{c|}{}                      &                                                                            & AC-Norm (Ours)                   & \textbf{78.52} & \textbf{78.18} & \textbf{82.84}      & \textbf{82.63}     \\ \cline{2-7} 
\multicolumn{1}{c|}{}                      & \multirow{4}{*}{BYOL}                                                      & Vanilla-FT                       & 76.84          & 77.52          & 77.48               & 77.33              \\
\multicolumn{1}{c|}{}                      &                                                                            & StochNorm~\cite{kou2020sn}                    & 75.38          & 75.76          & 77.92               & 77.81              \\
\multicolumn{1}{c|}{}                      &                                                                            & BSS~\cite{chen2019bss}                         & 76.35          & 78.44          & 78.87               & 80.23              \\ \cline{3-7} 
\multicolumn{1}{c|}{}                      &                                                                            & AC-Norm (Ours)                             & \textbf{77.95} & \textbf{79.00} & \textbf{79.84}      & \textbf{82.86}     \\ \hline
\multicolumn{1}{c|}{\multirow{4}{*}{Fully supervised}}  & \multirow{4}{*}{\begin{tabular}[c]{@{}c@{}}VFS $\rightarrow$ DFC\\ DFC $\rightarrow$ VFS\end{tabular}} & Vanilla-FT                       & 76.18          & 78.27          & 81.85               & 82.51              \\
\multicolumn{1}{c|}{}                      &                                                                            & StochNorm~\cite{kou2020sn}                         & 75.15          & 75.75          & 81.07               & 86.41              \\
\multicolumn{1}{c|}{}                      &                                                                            & BSS~\cite{chen2019bss}                           & 77.22          & 77.30          & 81.65               & 81.70              \\ \cline{3-7} 
\multicolumn{1}{c|}{}                      &                                                                            & AC-Norm  (Ours)                    & \textbf{78.01} & \textbf{78.72} & \textbf{82.34}      & \textbf{86.48}     \\ \hline
\multicolumn{2}{c|}{\multirow{3}{*}{Average gain over Vanilla-FT}}                                                      & \multicolumn{1}{l|}{Stoch-Norm~\cite{kou2020sn}} 
& $\downarrow$1.18 
& $\downarrow$0.69
& $\downarrow$0.35    
& {\color{blue}{$\uparrow$1.05} }\\
\multicolumn{2}{c|}{}                                                                                                   & \multicolumn{1}{l|}{BSS~\cite{chen2019bss}}          & $\downarrow$0.35   & \color{blue}{$\uparrow$0.06}    & \color{blue}{$\uparrow$0.03}   & {$\downarrow$0.32} \\ \cline{3-7} 
\multicolumn{2}{c|}{}                                                                                                   & \multicolumn{1}{l|}{AC-Norm (Ours)}     & \textbf{\color{blue}{$\uparrow$1.09}}   & \textbf{\color{blue}{$\uparrow$0.85}}   & \textbf{\color{blue}{$\uparrow$1.33}}   & \textbf{\color{blue}{$\uparrow$3.44}} \\ \hline
\end{tabular}}
\end{table}

\begin{table*}[]
\centering
\caption{Quantitative comparisons in various TL settings on 3D CT datasets. The best results are in bold and the positive average gain of each FT method against the baseline is highlighted in blue.}\label{tab::results_ct}
\resizebox{0.76\textwidth}{!}{
\begin{tabular}{cc|l|cc|ccc|ccc|cc}
\hline
\multicolumn{2}{c|}{\multirow{2}{*}{PT Method}}        & \multirow{2}{*}{FT Method} & \multicolumn{2}{c|}{NCS }        & \multicolumn{3}{c|}{LCS (DSC$\uparrow$ )} & \multicolumn{3}{c|}{LCS (HD95$\downarrow$ )} & \multicolumn{2}{c}{NCC}                                                 \bigstrut\\ \cline{4-13} 
                                     &  &                     & DSC$\uparrow$           & mIoU$\uparrow$            & Liver                   & Tumor                   & Avg.                   & Liver                  & Tumor                   & Avg.                   & AUC$\uparrow$                             & ACC$\uparrow$                              \bigstrut\\ \hline
                                   \multicolumn{1}{c|}{\multirow{16}{*}{Self-supervised}} &
\multirow{4}{*}{RPL}                   & Vanilla-FT          & 76.10          & 81.36          & 94.55                   & 65.38                   & 79.97                  & 3.57                   & 22.33                   & 12.95                  & 99.29                              & 97.87                              \\
                                  \multicolumn{1}{c|}{} &    & StochNorm~\cite{kou2020sn}           & 75.22          & 81.58          & 94.84                   & 60.13                   & 77.49                  & 7.49                   & 24.87                   & 16.18                  & 99.36                              & 97.11                              \\
                                  \multicolumn{1}{c|}{}    &       & BSS~\cite{chen2019bss}                 & 75.39          & 81.72          & 94.39                   & 66.28                   & 80.34                  & \textbf{3.18}                   & 24.52                   & 13.85                  & 98.72                              & 96.97                              \\ \cline{3-13} 
                                    \multicolumn{1}{c|}{}    &      & AC-Norm (Ours)            & \textbf{76.53} & \textbf{82.24} & \textbf{94.91}          & \textbf{71.45}          & \textbf{83.18}         & {4.22}          & \textbf{18.94}          & \textbf{11.58}         & \textbf{99.37}                     & \textbf{98.85}                     \\ \cline{2-13}
  \multicolumn{1}{c|}{} &    \multirow{4}{*}{RKB+}                  & Vanilla-FT          & 74.31          & 80.82          & 94.64                   & 60.02                   & 77.33                  & 3.78                   & 23.15                   & 13.47                  & 98.88                              & \multicolumn{1}{c}{98.48}          \\
                                \multicolumn{1}{c|}{}     &        & StochNorm~\cite{kou2020sn}           & 72.73          & 80.52          & 94.06                   & 64.58                   & 79.32                  & 4.12                   & 18.53                   & 11.33                  & 98.62                              & \multicolumn{1}{c}{97.74}          \\
                                \multicolumn{1}{c|}{}   &           & BSS~\cite{chen2019bss}                 & 73.73          & 80.37          & 94.48                   & 64.90                   & 79.69                  & 3.26                   & 18.60                   & 10.92                  & 98.55                              & \multicolumn{1}{c}{98.22}          \\ \cline{3-13} 
                                    \multicolumn{1}{c|}{}   &       & AC-Norm (Ours)             & \textbf{75.29}          & \textbf{80.97}          & \textbf{95.31}          & \textbf{67.46}          & \textbf{81.39}         & \textbf{2.32}          & \textbf{17.73}          & \textbf{10.02}         & \textbf{99.27}                     & \multicolumn{1}{c}{\textbf{99.01}} \\ \cline{2-13}
  \multicolumn{1}{c|}{} &    \multirow{4}{*}{PCRL}                  & Vanilla-FT          & 75.60          & 81.97         & 93.76                  & 64.08                  & 78.92
                  & 19.18                  & 18.44                   & 18.81                  & 98.97                              & \multicolumn{1}{c}{97.88}          \\
                           \multicolumn{1}{c|}{}     &              & StochNorm~\cite{kou2020sn}           & 73.10          & 80.84          & 94.02                   & 61.51                   & 77.77                  & 9.29                   & 24.47                   & 16.88                  & 99.34                              & \multicolumn{1}{c}{98.14}          \\
                            \multicolumn{1}{c|}{}    &              & BSS~\cite{chen2019bss}                 & 74.58 & 81.34          & \textbf{95.03}                        &  62.96
                       &   79.00
                      &              \textbf{3.23}          & 19.11                        &11.17                        & 98.56                              & \multicolumn{1}{c}{99.17}          \\ \cline{3-13} 
                            \multicolumn{1}{c|}{}      &            & AC-Norm (Ours)            & \textbf{75.68}          & \textbf{82.04} & 94.71         & \textbf{67.73}          & \textbf{81.22}         & 3.45          & \textbf{16.90}          & \textbf{10.18}         & \textbf{99.52}                             & \multicolumn{1}{c}{\textbf{99.45}}          \\ \cline{2-13}
  \multicolumn{1}{c|}{} &    \multirow{4}{*}{SimCLR}                & Vanilla-FT          & 75.96          & 82.00          & 93.91                   & 60.24                   & 77.08                  &     5.63                   &      25.27                   &   15.45                     & \multicolumn{1}{c}{99.29}          & \multicolumn{1}{c}{97.91}          \\
                                 \multicolumn{1}{c|}{}     &        & StochNorm~\cite{kou2020sn}           & 74.85          & 81.32          & 94.55          & 61.27                   & 77.95                  & 4.77                   & 24.65                   & 14.71                  & \multicolumn{1}{c}{\textbf{99.46}} & \multicolumn{1}{c}{97.43}          \\
                                 \multicolumn{1}{c|}{}    &         & BSS~\cite{chen2019bss}                 & 76.02          & 82.56          & 83.08                   & 54.73                   & 68.91                  &       22.78                 &   30.45                      &         26.61               & \multicolumn{1}{c}{99.32}          & \multicolumn{1}{c}{97.18}          \\ \cline{3-13} 
                                  \multicolumn{1}{c|}{}    &        & AC-Norm (Ours)            & \textbf{77.16} & \textbf{82.78} &  \textbf{94.58}                   & \textbf{62.62}                   & \textbf{78.60}         &  \textbf{3.74}                     &  \textbf{24.09}                       &   \textbf{13.91}                     & \multicolumn{1}{c}{99.44}          & \multicolumn{1}{c}{\textbf{98.70}} 
                                  \\ \hline

  
  \multicolumn{1}{l|}{\multirow{8}{*}{Fully-supervised}} & \multicolumn{1}{c|}{\multirow{4}{*}{\begin{tabular}[c]{@{}c@{}}LCS $\rightarrow$ NCS\\ NCS $\rightarrow$ LCS\\NCS $\rightarrow$ NCC\end{tabular}}}

 & Vanilla-FT          & 73.42          & 80.41          & 94.86                   & 59.53                   & 77.20                  &                  \textbf{2.73}      &   22.16                      &                     12.45   & 98.32                              & 93.73                              \\
                    \multicolumn{1}{l|}{}   &        \multicolumn{1}{c|}{}          & StochNorm~\cite{kou2020sn}          & 72.41          & 80.19          & 94.42                   & 60.65                   & 77.54                  &    3.88                    &    24.56                     &         14.22               & 99.15                              & 94.05                              \\
                             \multicolumn{1}{l|}{}   &          \multicolumn{1}{c|}{}     & BSS~\cite{chen2019bss}                & 74.81          & 80.60          & 93.89                   & 60.97                   & 77.43                  &             4.04           &   25.99                      &  15.02                      & 98.35                              & 94.58                            \\ \cline{3-13} 
                               \multicolumn{1}{l|}{}   &        \multicolumn{1}{c|}{}     & AC-Norm  (Ours)           & \textbf{74.83} & \textbf{81.02} & \textbf{95.12}          & \textbf{61.37}          & \textbf{78.25}         &              2.98          &   \textbf{18.37} &    \textbf{10.68}                    & \textbf{99.50}                     & \textbf{98.58}                     \\ \cline{2-13}
 \multicolumn{1}{l|}{}   & \multicolumn{1}{c|}{\multirow{4}{*}{\begin{tabular}[c]{@{}c@{}}NCC $\rightarrow$ {NCS} \\ NCC $\rightarrow$ LCS\\LCS $\rightarrow$ NCC\end{tabular}}}& Vanilla-FT          & 75.16          & 81.73          & 93.46                   & 62.15                   & 77.81                  &   5.12                     &                  20.64       &     12.88                   & 99.26                              & 98.48                              \\
                               \multicolumn{1}{l|}{}   &          & StochNorm~\cite{kou2020sn}           & 74.99          & 81.35          & 94.43                   & 60.54                   & 77.49                  & 4.74                       &  21.55                       &  13.15                      & 99.07                              & \textbf{99.06}                     \\
                                  \multicolumn{1}{l|}{}   &      & BSS~\cite{chen2019bss}                 & 75.63          & 81.78          & 94.43                   & 64.96                   & 79.76                  &   3.48                   &  20.42              &   11.95                   & 99.39                              & 98.77                              \\ \cline{3-13} 
                                \multicolumn{1}{l|}{}   &        & AC-Norm (Ours)            & \textbf{75.96} & \textbf{81.93} & \textbf{95.27}          & \textbf{69.00}          & \textbf{82.14}         &    \textbf{2.49}                     &  \textbf{16.62}                        &    \textbf{9.56}                     & \textbf{99.51}                     & 99.02                              \\ \hline

\multicolumn{2}{c|}{\multirow{3}{*}{Average gain over Vanilla-FT}} & StochNorm~\cite{kou2020sn} & $\downarrow$2.03 & $\downarrow$0.42 & \color{blue}{$\uparrow$0.19} & $\downarrow$0.45 & $\downarrow$0.13 & \color{blue}{$\downarrow$0.95} & $\uparrow$1.11 & $\uparrow$0.08 & \color{blue}{$\uparrow$0.17} & $\downarrow$0.14 \\
\multicolumn{2}{c|}{}                          & BSS~\cite{chen2019bss}       & $\downarrow$0.07 & $\downarrow$0.01 & $\downarrow$1.65 & \color{blue}{$\uparrow$0.57} & $\downarrow$0.54 & \color{blue}{$\downarrow$0.01} & $\uparrow$1.18 & $\uparrow$0.59 & $\downarrow$0.19 & \color{blue}{$\uparrow$0.09} \\ \cline{3-13} 
\multicolumn{2}{c|}{}                          & AC-Norm (Ours)   & \textbf{\color{blue}{$\uparrow$0.82}} & \textbf{\color{blue}{$\uparrow$0.45}} & \textbf{\color{blue}{$\uparrow$0.79}}& \textbf{\color{blue}{$\uparrow$4.71}} & \textbf{\color{blue}{$\uparrow$2.75}} & \textbf{\color{blue}{$\downarrow$3.47}} & \textbf{\color{blue}{$\downarrow$3.22}} & \textbf{\color{blue}{$\downarrow$3.35}} & \textbf{\color{blue}{$\uparrow$0.43}} & \textbf{\color{blue}{$\uparrow$1.54}} \\ \hline
\end{tabular}}
\end{table*}
The distribution gap between LCS and 3D-MG is larger than NCS and NCC as the 3D pretraining data captured in the lungs is distinct from the abdomen in LCS. As a reconstruction task, MG concentrates more on pixel-wise information which is preferable for the segmentation over the classification task. Then, we can assume that the gap in 3D-MG$\rightarrow$LCS and 
3D-MG$\rightarrow$NCC are larger in 
3D-MG$\rightarrow$NCS.
We can observe that some UDA techniques, which are effective in 3D-MG$\rightarrow$NCS, fail in 3D-MG$\rightarrow$LCS/NCC with a larger domain gap. In particular, some methods based on mixed BN statistics have a high risk of collapse, e.g. $\alpha$-BN, Auto-DIAL in 3D-MG$\rightarrow$LCS/NCC
and 2D-MG$\rightarrow$DFC.
This indicates that the forced mixing of statistics ($\alpha$-BN, Auto-DIAL) or over-reliance on statistics (TransNorm) might have a negative effect on the transfer performance under a significant domain gap. Different from such normalization techniques that only consider the statistics, the proposed AC-Norm boost the transfer efficiency by importance-aware finetuning based on affine parameters, leading to the best performance. 
\subsection{Results in diverse TL settings.}
In addition to MG pretraining, we perform AC-Norm finetuning on transferring diverse pretrained models to target tasks. The results of 2D fundus image target tasks and 3D CT image target tasks are respectively reported in Table~\ref{tab::results_fundus}
 and Table~\ref{tab::results_ct}. In DFC, it can be observed that AC-Norm consistently achieves the best KAPPA. Although StochNorm attains the highest accuracy in Jigsaw$\rightarrow$DFC, its poor KAPPA indicates that most of the cases are incorrectly predicted as the major class.
The main challenges of VFS are the connectivity of vessel segmentation and recall of small vessels that are prone to be missed. Thus, we focus on the CL-DSC~\cite{paetzold2019cldice} and Sensitivity metrics. The results prove that AC-Norm finetuning contributes to more accurate vessel segmentation.
Table~\ref{tab::results_ct} indicates that AC-Norm significantly surpasses StochNorm and BSS across all transfer settings with two exceptions: transfer under a small domain gap and the performance saturation in NCC. Particularly, the domain gaps in NCC$\xrightarrow{}$NCS and LCS$\xrightarrow{}$NCS, are relatively smaller due to the same training data of NCC and NCS and the same segmentation task of LCS and NCS. In these scenarios, BSS achieves close performance to ours. However, BSS and StochNorm yield marginal gains or even fail to improve the baseline in most TL setups.

\subsection{AC-Corr as a criterion for pretrained model selection}
\begin{figure}[]
\centering
\includegraphics[width=0.85\linewidth]{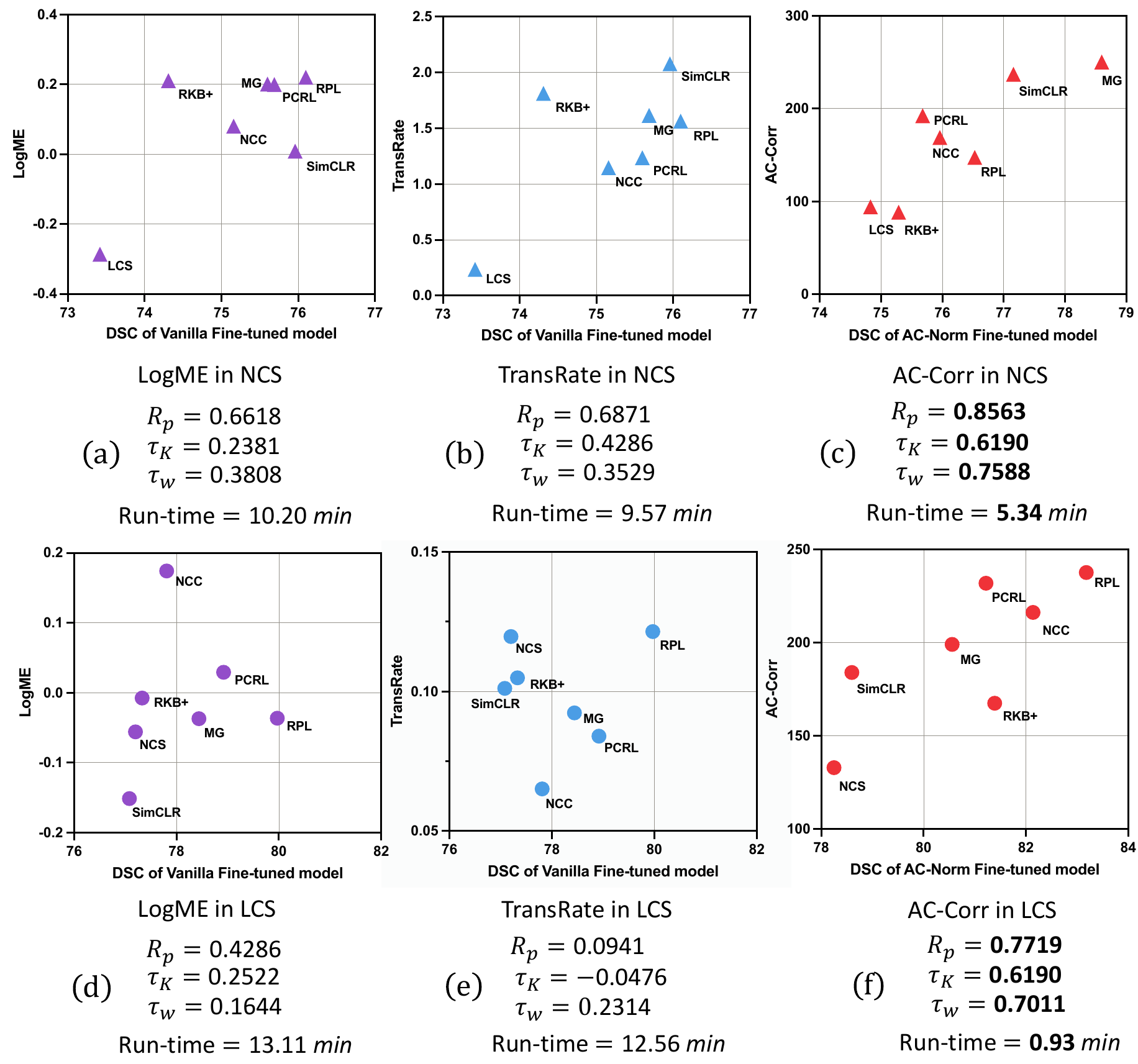}
\caption{Transferability estimation in 7 pretrained models (both FSL and SSL) to NCS and LCS. $R_p$, $\tau_K$ and $\tau_w$ denote the Pearson correlation coefficient, the Kendall’s $\tau$ and the weighted $\tau$. The run time (minutes) for each method to assess one pretrained model is reported.}
\label{fig::TE}
\end{figure}

Here we evaluate the performance of AC-Corr and other baseline measures for assessing the transferability of multiple pretrained models to a specific target task. The target performance of a pretrained model after the entire finetuning phase is naturally the ground truth of its transferability. It is worth noting that LogME~\cite{you2021logme} and TransRate~\cite{huang2022frustratingly} relying on the static pretrained representations are designed to estimate the performance of the vanilla fintuned model. AC-Corr considers cross-channel transfer within one-epoch finetuning so it is supposed to estimate the performance of AC-Norm instead of the vanilla fintuned model. Following related works~\cite{you2021logme, huang2022frustratingly}, we compute the Pearson correlation coefficient, 
the Kendall’s $\tau$~\cite{kendall1938new} and the weighted $\tau$ to measure the performance of TE methods. We compute the above metrics between the rankings of the TE scores and the test performance in the target task. Results shown in Fig.~\ref{fig::TE} confirm that AC-Corr predicts the ranking of 7 pretrained models on both NCS and LCS more correctly than other baselines and costs less run time even though it requires one-epoch tuning. 

\section{Discussion}

\begin{table}[t]
\renewcommand{\arraystretch}{1}
\centering
\caption{Results of transferring the MG pretrained model to NCS and LCS with different variations of the proposed AC-Norm.}\label{tab::Ablation}
\resizebox{0.48\textwidth}{!}{
\begin{tabular}{cc|cc|ccc}
\hline
\multicolumn{2}{c|}{\multirow{2}{*}{Method}}                            & \multicolumn{2}{c|}{NCS}        & \multicolumn{3}{c}{LCS (DSC$\uparrow$)}                          \\ \cline{3-7} 
\multicolumn{2}{c|}{}                                                   & DSC$\uparrow$             & mIoU$\uparrow$            & Liver          & Tumor          & Avg.           \\ \hline
\multicolumn{1}{c|}{\multirow{6}{*}{Full FT}}    & MG-Vanilla-FT             & 75.69          & 82.13          & 94.03          & 62.85          & 78.44          \\
\multicolumn{1}{c|}{}                            & +SC-Norm             & 76.52          & 82.10          & 94.21          & 60.13          & 77.17          \\
\multicolumn{1}{c|}{}                            & +AC-Norm w/ $C_{diag}$   & 76.88          & 82.26          & 93.66          & 64.87          & 79.27          \\
\multicolumn{1}{c|}{}                            & +AC-Norm w/ $C_{non-sparse}$ & 76.34          & 81.51          & 94.89          & 64.58
          & 79.74          \\
\multicolumn{1}{c|}{}                            & +AC-Norm w/ $C_{opt}$   & 77.15          & 82.08          & 92.86          & 61.01          & 76.94          \\
\multicolumn{1}{c|}{}                            & +AC-Norm             & \textbf{78.60} & \textbf{82.87} & \textbf{95.14} & \textbf{65.84} & \textbf{80.56} \\ \hline
\multicolumn{1}{c|}{\multirow{2}{*}{Partial FT}} & Training only BN     & 72.51          & 78.33          & 89.31          & 55.80          & 72.56          \\
\multicolumn{1}{c|}{}                            & Training only AC-Norm    & \textbf{74.47} & \textbf{78.82} & \textbf{92.70} & \textbf{63.27} & \textbf{77.99} \\ \hline
\end{tabular}}
\end{table}
\subsection{Ablation study}
We conduct ablation studies in Table~\ref{tab::Ablation} to investigate the impact of different variants of recalibration matrix $C$.

First, we utilize BN statistics to replace  BN affine parameters as the indicator of domain information. In specific, $z^{s, t} = \frac{\mu^{s, t}}{\sqrt{({\sigma^{s,t}})^2+\epsilon}}$ is adopted to reflect the distribution similarity of source and target features per channel (named as SC-Norm), where $\{\mu_s, \sigma_s\}$ are the pretrained moving statistics and $\{\mu_t, \sigma_t\}$ are the BN statistics (mini-batch statistics for training and moving statistics for inference) in the target model. 
We can observe that the proposed AC-Norm exceeds the SC-norm, showing the superiority of affine parameters in representing the domain distributions. This insight is also supported by the observations in Fig.~\ref{fig::affine_sta_kernel}.
Next, we employ a diagonal $C_{diag}$ for AC-Norm without taking cross-channel correlations into account and a non-sparse $C_{non-sparse}$ without the sparsity operation in Eq.\ref{eq::sparsity}. The inferior performance of $C_{diag}$ and $C_{non-sparse}$ compared to cross-channel $C$ respectively validates the importance of cross-channel alignment and attention sparsity. In addition, we set $C$ to be fully trainable by network optimization, which we believe is an essential baseline. The inferior performance of trainable $C_{opt}$ suggests that simply introducing extra parameters lacks explicit constraints. In contrast, harnessing the information from the source model is valuable for transfer.

Motivated by the promising performance of partial finetuning in \cite{kanavati2021partial}, we fintune only BN layers in the target model with the remaining weights being fixed. The results show that AC-Norm can significantly improve the transfer performance over the vanilla BN when only partial parameters are trainable, which shows its potential in parameter-efficient tuning.

\subsection{Cross-modality transfer}

\begin{table}[]
\centering
\caption{Performance on transferring lung-CT-pretrained models to the MRI cardiac segmentation task CMS. The objects of interest in CMS comprise the left ventricle (LV), right ventricle (RV) and myocardium (MYO).}\label{tab::CMS}
\resizebox{0.48\textwidth}{!}{
\begin{tabular}{c|c|p{0.5cm}<{\centering}p{0.5cm}<{\centering}p{0.5cm}<{\centering}p{0.5cm}<{\centering}|p{0.5cm}<{\centering}p{0.5cm}<{\centering}p{0.5cm}<{\centering}p{0.5cm}<{\centering}}
\hline
\multirow{2}{*}{PT Method}   & \multirow{2}{*}{FT Method} & \multicolumn{4}{c|}{DSC$\uparrow$}                                     & \multicolumn{4}{c}{HD95$\downarrow$}                                  \\ \cline{3-10} 
                      &                     & LV             & RV             & MYO            & Avg.           & LV            & RV             & MYO           & Avg.          \\ \hline
Random Init.                    & From-scratch         & 91.42          & 82.71          & 84.48          & 86.20          & 16.04         & 24.82          & 5.59          & 15.48         \\ \hline
\multirow{4}{*}{MG}   & Vanilla-FT          & 93.30          & 86.18          & 85.84          & 88.44          & 3.44          & 10.27          & 4.25          & 8.98          \\
                      & StochNorm           & 92.19          & 84.54          & 85.80          & 87.51          & 2.62          & 8.76           & 4.76          & 5.38          \\
                      & BSS                 & 94.57          & 86.13          & 86.88          & 89.19          & 1.99          & 13.22          & 9.85          & 8.35          \\ \cline{2-10} 
                      & AC-Norm             & \textbf{94.31} & \textbf{87.64} & \textbf{87.19} & \textbf{89.71} & \textbf{2.16} & \textbf{7.52}  & \textbf{2.38} & \textbf{4.02} \\ \hline
\multirow{4}{*}{PCRL} & Vanilla-FT          & 88.79          & \textbf{88.36} & 86.82          & 87.99          & 17.50         & 18.36          & 4.72          & 13.53         \\
                      & StochNorm           & 93.85          & 83.98          & 85.33          & 87.72          & 2.32          & 22.22          & 2.67          & 9.07          \\
                      & BSS                 & 93.59          & 85.44          & 85.87          & 88.30          & 7.79          & 10.34          & 2.43          & 6.85          \\ \cline{2-10} 
                      & AC-Norm             & \textbf{94.19} & 86.52          & \textbf{87.01} & \textbf{89.24} & \textbf{2.39} & \textbf{11.65} & \textbf{2.15} & \textbf{5.39} \\ \hline
\multirow{4}{*}{RKB+} & Vanilla-FT          & 94.49          & 85.51          & 87.40          & 89.10          & 2.05          & 18.01          & 2.08         & 7.38         \\
                      & StochNorm           & 94.56          & 85.39          & 86.76          & 88.90          & 2.05          & 17.79          & 2.36          & 7.40          \\
                      & BSS                 & 93.45          & 86.48          & 86.92          & 88.95          & 2.47          & 12.43          & 2.44          & 5.78          \\ \cline{2-10} 
                      & AC-Norm             & \textbf{94.62} & \textbf{86.84} & \textbf{87.38} & \textbf{89.61} & \textbf{1.99} & \textbf{10.57} & \textbf{2.20} & \textbf{4.92} \\ \hline

\end{tabular}}
\end{table}

In this section, we perform AC-Norm in a cross-modality transfer from MG, PCRL, and RKB+ that are pretrained on lung CT images to the MRI cardiac segmentation task CMS. From Table~\ref{tab::CMS}, we can see that cross-modality pretraining still brings remarkable gains ($>$2\%) over random initialization in CMS, even under significant domain shifts. With channel recalibration, AC-Norm exceeds previous finetuning approaches in the average DSC and HD95, proving its generic effectiveness in diverse transfer learning scenarios. 

\subsection{Visualizations}
\begin{figure}[!t]
\centering
\includegraphics[width=0.9\linewidth]{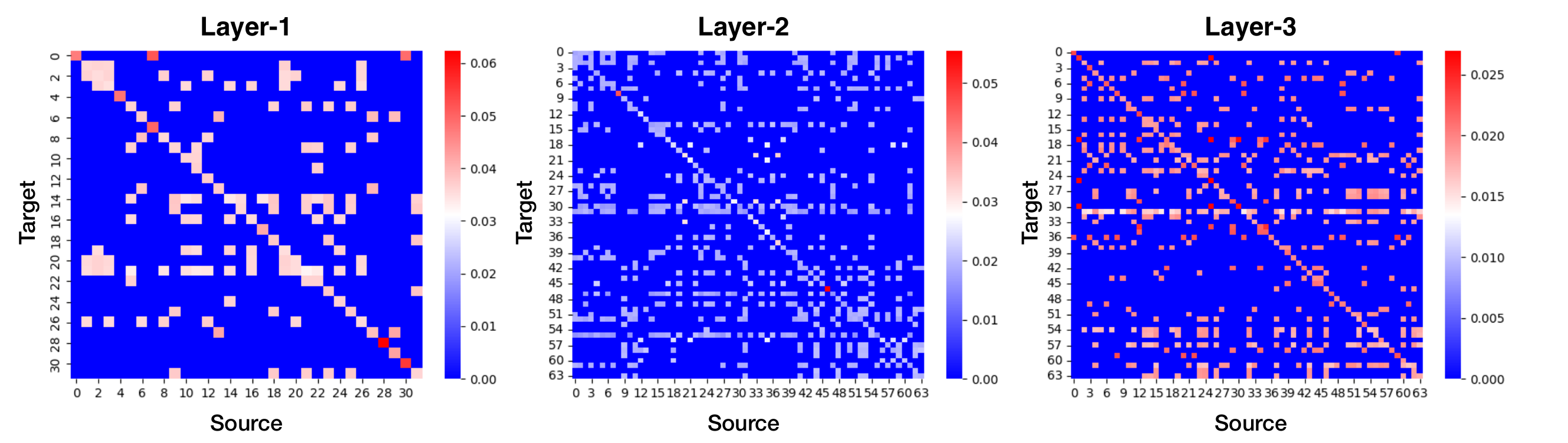}
\caption{Heatmaps of the cross-channel correlation matrix $C$ in the AC-Norm for MG $\rightarrow$ NCS from the first to the third layer in the encoder.}
\label{fig4}
\end{figure}
Towards a better understanding of AC-Norm, we visualize the calibration matrix $C$ in MG $\rightarrow$ NCS.
As illustrated in Fig.~\ref{fig4}, AC-Norm adaptively modulates the attention values for BN channels, facilitating sufficient knowledge transfer. Each value in the heatmap represents the transferability of a source channel to a target channel. We notice that the transferability varies in different layers, which is in accordance with \cite{amiri2020fine}.

\begin{figure}[]
\centering
\includegraphics[width=0.95\linewidth]{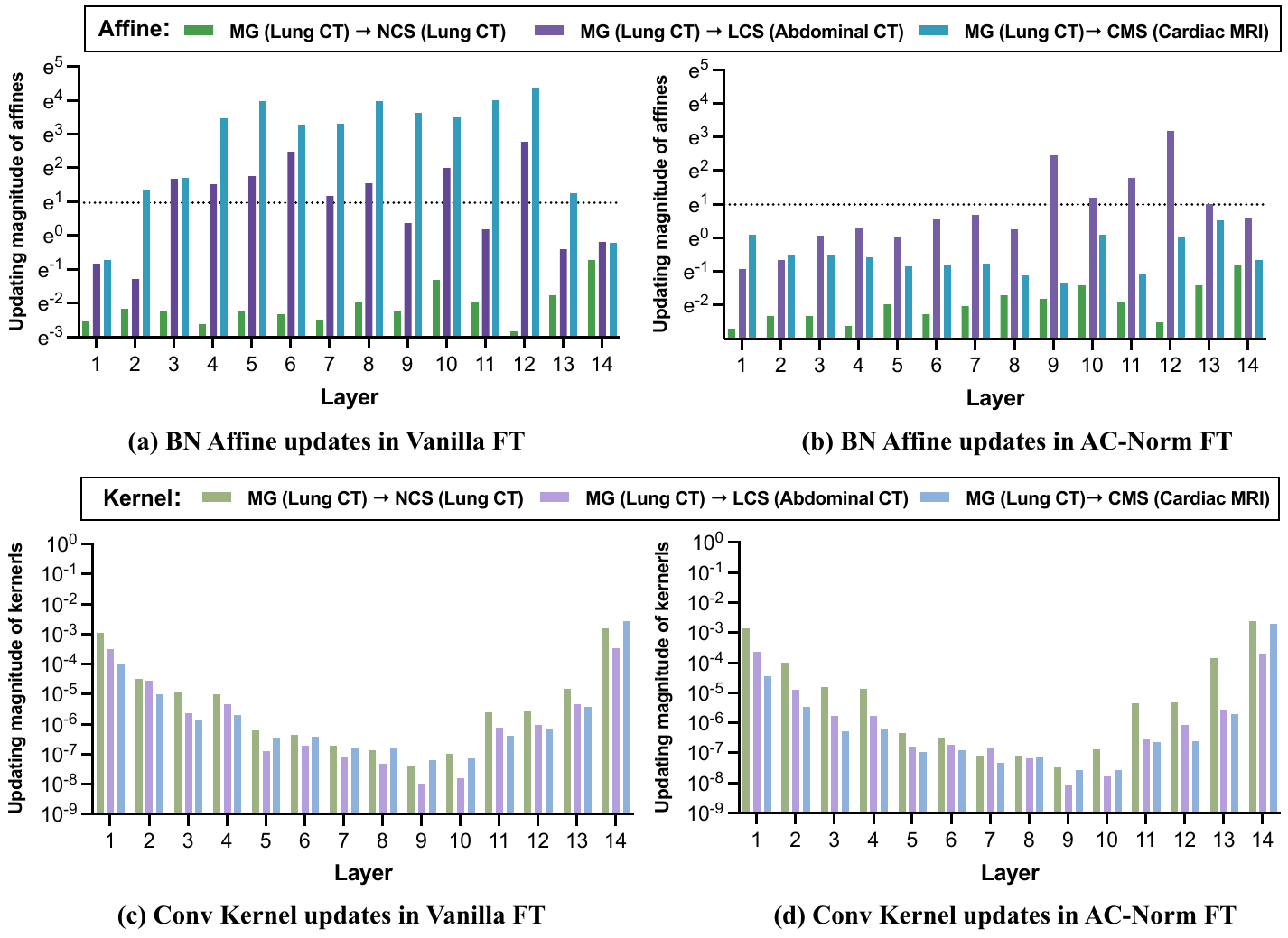}
\caption{The updating magnitudes of affine and kernel parameters in vanilla FT and AC-Norm FT at different layers.}
\label{fig::affine_per_layer}
\end{figure}

To further understand how AC-Norm assists in tackling domain shifts, we display the updating magnitudes in BN affine and convolutional kernel parameters at different layers. First of all, the comparison of Fig.~\ref{fig::affine_per_layer}(a)(b) and (c)(d) indicates that affine parameters dominate the updates in network parameters while kernel parameters barely change. It can be observed from Fig.~\ref{fig::affine_per_layer}(a) that the affine parameters update dramatically in MG $\rightarrow$ LCS/CMS at most of the layers due to the notable gaps between lung CT data to abdominal CT/cardiac MRI data, highlighting low reusability of pretrained weights. Fig.~\ref{fig::affine_per_layer}(b) implies that using AC-Norm increases the reusability of pretrained weights in lower and middle layers for MG $\rightarrow$ LCS/CMS, which accounts for its better transfer performance. Intriguingly, we notice that the updates of convolutional kernel parameters also appear as a U-shape. This phenomenon can be explained by the difficulty of propagating gradients to bottleneck layers in U-Net.
\section{Conclusion}
In this paper, without adding complex modules, we present AC-Norm as a new finetuning approach to further improve transfer efficiency when using pretrained models to compensate for the shortage of medical supervision in downstream clinical applications. AC-Norm recalibrates target BN channels by formulating the distribution similarity of source and target affine-transformed features and assigning transferability-aware importance values to different channels. Thorough experiments demonstrate the superiority of AC-Norm across multiple pretrained models and medical datasets against other state-of-the-art finetuning methods. Moreover, AC-Norm shows a promising benefit in transferability assessment. One limitation is that AC-Norm can only be currently used in architectures equipped with BN layers. In a concurrent work, introducing additional affine transformations on features has showcased remarkable performance in finetuning Transformer-based foundation models as well~\cite{lian2022scaling}. We expect to extend AC-Norm to more architectures without BN layers in future work.


\nocite{*}
\bibliographystyle{IEEEtran}
\bibliography{reference}

\end{document}